\documentclass[conference]{IEEEtran}
\ifCLASSINFOpdf
\else
\fi
\usepackage{amsmath}
\usepackage{multirow}
\usepackage{colortbl}
\usepackage{graphicx}
\usepackage{caption}
\usepackage{subfig}
\usepackage{blindtext}
\usepackage{float}

\begin{document}
%
\title{Efficient Eye Typing with 9-direction Gaze Estimation}

\author{\IEEEauthorblockN{Chi Zhang}
\IEEEauthorblockA{School of Computer Science and \\Technology\\
China University of \\Mining and Technology\\
Xuzhou, 221116, China\\
Email: chizhang@cumt.edu.cn}
\and
\IEEEauthorblockN{Rui Yao}
\IEEEauthorblockA{School of Computer Science and \\Technology\\
China University of \\Mining and Technology\\
Xuzhou, 221116, China\\
Email: ruiyao@cumt.edu.cn}
\and
\IEEEauthorblockN{Jinpeng Cai}
\IEEEauthorblockA{College of Engineering \& \\Computer Science\\
 Australian National University\\
 Canberra, 2601, Australia\\
Email: u5715685@anu.edu.au}}


%


\maketitle

\begin{abstract}
Vision based text entry systems aim to help disabled people achieve text communication using eye movement. Most previous methods have employed an existing eye tracker to predict gaze direction and design an input method based upon that. However, these methods can result in eye tracking quality becoming easily affected by various factors and lengthy amounts of time for calibration. Our paper presents a novel efficient gaze based text input method, which has the advantage of low cost and robustness. Users can type in words by looking at an on-screen keyboard and blinking. Rather than estimate gaze angles directly to track eyes, we introduce a method that divides the human gaze into nine directions. This method can effectively improve the accuracy of making a selection by gaze and blinks. We build a Convolutional  Neural Network (CNN) model for 9-direction gaze estimation. On the basis of the 9-direction gaze, we use a nine-key T9 input method which is widely used in candy bar phones. Bar phones were very popular in the world decades ago and have cultivated strong user habits and language models. To train a robust gaze estimator, we created a large-scale dataset with images of eyes sourced from 25 people. According to the results from our experiments, our CNN model is able to accurately estimate different people's gaze under various lighting conditions by different devices. In considering disable people¡¯s needs, we removed the complex calibration process. The input methods  can run in screen mode and portable off-screen mode. Moreover, The datasets used in our experiments are made available to the community to allow further experimentation.
\end{abstract}


%
\IEEEpeerreviewmaketitle

\section{Introduction}
\label{sec:intro}

Vision based human-computer interaction (HCI) is a hot topic in the field of computer vision. We design a low-cost gaze controlled text entry method to help disabled people those who may be unable to carry out the basic activities of daily living but can move their eyes achieve text communication.

\begin{figure}[!h]
\centering
 \includegraphics[width=0.5\textwidth]{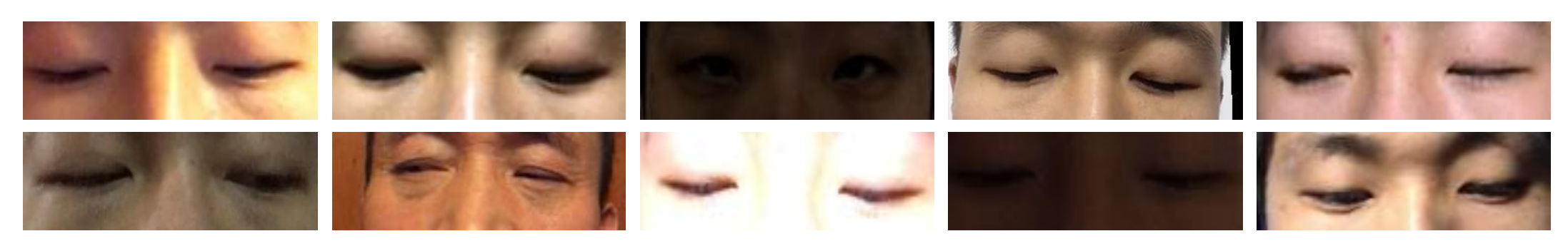}
\caption{Eye images that are difficult to be used for estimating gaze}
\vspace{-2mm}
\label{fig:01}
\end{figure}

A typical video-based eye typing system uses a camera to capture eye movement, from which users can type in words by looking at an on-screen keyboard. This process can be divided into two parts: the first part is an eye tracking system and the other part is a text entry system. The eye tracking system aims to detect eyes from images and then estimate the gaze direction. Before using the system, users must do the calibration to map estimated directions to corresponding positions on the screen. For the text input system, it is based on where people are looking at on the screen to make selections and input text. In previous works, researchers usually design systems on the basis of the existing eye trackers and they only need to design text input systems. However, the following issues widely exist in such systems: first, the quality of the eye tracker constrains the design of the text input system. A high accuracy eye tracking system can estimate human gaze with less error, so it can precisely select the targets even if there are more intensive objects on the screen. A high accuracy tracking system demands high precision equipment, with low precision equipment unable to meet the input system accuracy requirement. Accurate eye tracking requires high-resolution human eye images, which demands a high quality camera and optimal lighting conditions. This increases the cost and limits the range of applicable environments. Furthermore, if the device is changed, the camera probably needs to be recalibrated too. So, most previous systems are device-specific. Currently, there are still many challenges for eye tracking that need to be overcome: low-resolution images, poor lighting conditions and physiological responses (such as people looking down having a natural squint and tired people showing droopy eyelids). Fig.~\ref{fig:01} illustrates examples of eye images that are hard to use for estimating gaze. Secondly, the accuracy of human gaze angle changes is $0.5^{\circ}$ ~\cite{twentyyear}, and even when people stare in a fixed direction, there still exists a jittering movement of the eyes and the body also moves subconsciously, both of which can make it difficult to execute accurate and stable eye tracking. Finally, an eye tracker requires a lot of time to calibrate before using. During the calibration process, people are not allowed to move their body, and slight movements may result in an error in the final estimation. As usage time increases, the calibrated system can become skewed and users may need to calibrate the system again. It is easy to foresee how these inconveniences could become an issue for disabled people attempting to use these systems in their daily life.

\begin{figure}[t]
\centering
\includegraphics[width=.5\textwidth]{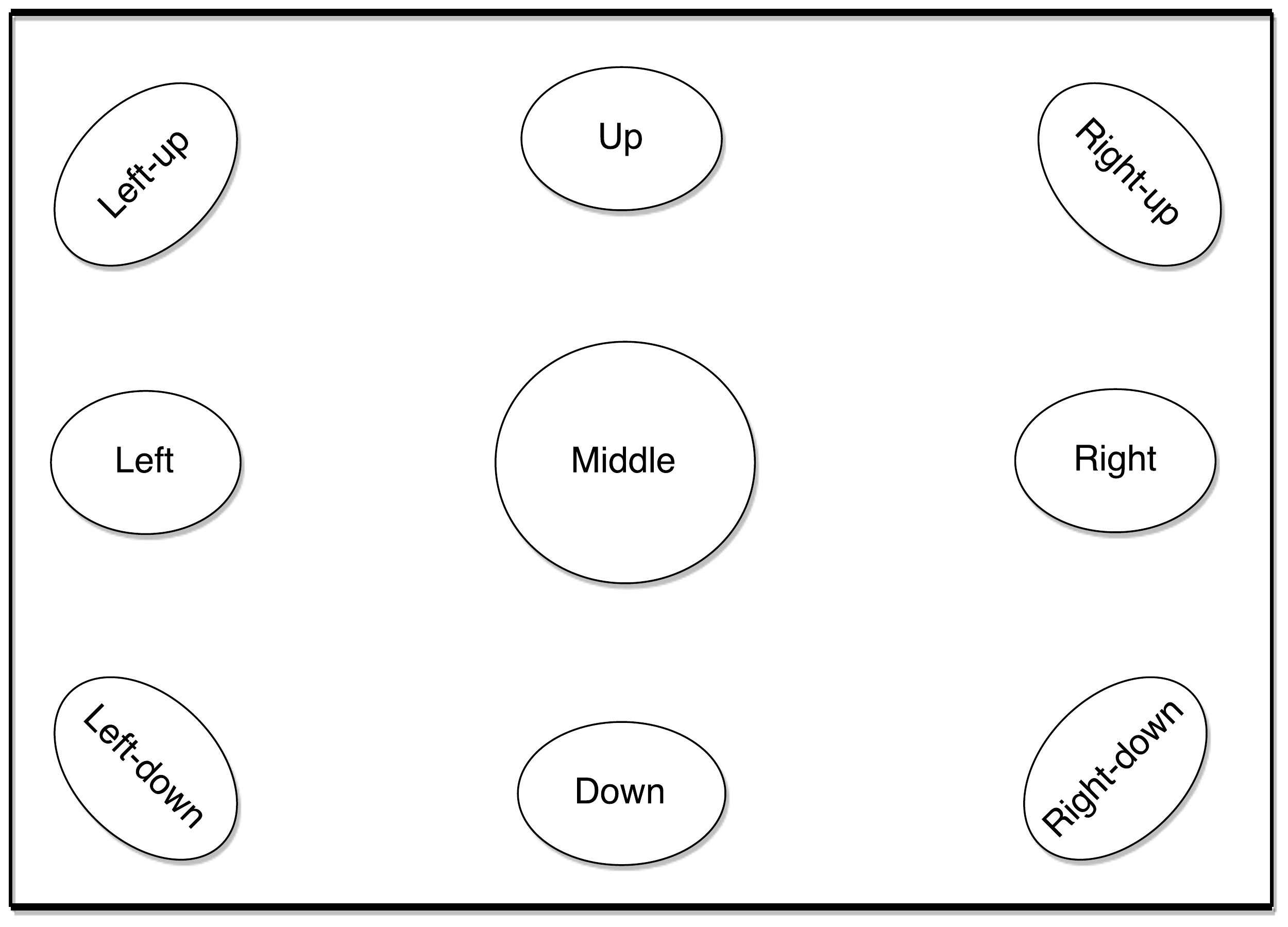}
\caption{9 Corresponding regions on the screen of the 9 directions}
\label{fig:02}
\end{figure}

In order to solve the problems above and consider the needs of disabled people in the application of these systems, instead of employing an existing eye tracker, we attempt to solve the eye tracking problem and input method problems together. We separate human gaze into nine directions: left-up, up, right-up, left, middle, right, left-down, down and right-down. The division has the following advantages: first, generally speaking the vertical visual field of both eyes is $100^{\circ}$ and the horizontal visual field is $200^{\circ}$~\cite{visualfield1,visualfield2}. If we divide those angles into nine directions, every direction has the range from $30^{\circ}$ to $70^{\circ}$. The angle range allows the gaze estimator to have a larger fault tolerant space, which means the nine directions can be recognized with lower error. This can also lower the required standard for the devices employed and the conditions it is used under. In other words, it reduces the cost and increases the range of applicable environments. Second, it can simplify the process of calibration. The calibration process usually contains camera-calibration, geometric-calibration, personal calibration and gaze mapping calibration~\cite{hansen2010eye}. The whole process aims to determine a bunch of parameters. A common calibration method is to project points on the screen and have the user follow the points with their eyes accordingly. The calibration process becomes easier if we simplify eye gaze into nine directions. As the nine directions are centrosymmetric, if we want to map nine directions of gaze to nine regions of the screen when using the system (see Fig.~\ref{fig:02}), we only need to keep the head stationary and eyes level with the center of screen. After adjusting the distance between the eyes and the screen, we can execute the calibration process easily. Moreover, due to the large fault tolerant space, slight body movement will not affect the calibration results. We introduce a CNN model to estimate human gaze. The estimation results of the CNN are ten states of eyes which include nine gaze directions and one eye-closed state. We train the CNN model to learn the mapping between the ten categories and corresponding eye images. A big advantage of using CNN to estimate gaze is that CNN methods do not require complex calibration to determine a series of parameters, and in some CNN based methods, calibration is even not needed~\cite{krafka2016eye}.

\begin{figure}[t]
\centering
 \includegraphics[width=.5\textwidth]{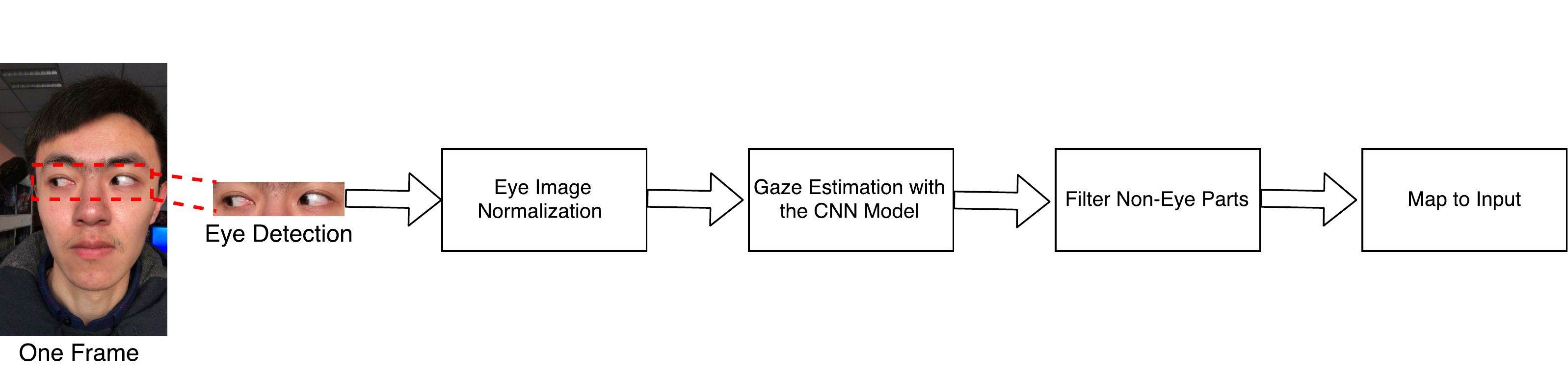}
\caption{Overview of our vision based text input method. For each frame of the video sequence, we first detect both eyes from the image. Then we normalize the eye images by resizing them to 32 $\times$ 128 pixels. We use the CNN model to estimate the states of the eyes and to eliminate false detected eyes. We input text according to the estimated results of each frame.}
\label{fig:07}
\end{figure}

When designing the text entry system, many factors need to be considered. Tradeoffs often must be made between accuracy, text entry speed, cost and accessibility. Our goal is to design a low-cost input method which can use off-the-shelf devices to help disabled people type in words with high accuracy and adequate speed. The text entry system can be used in daily life rather than just being limited to laboratory conditions. To achieve this goal, the methods needs to withstand influence from factors including false eye detection, involuntary blinks, saccade, illumination, and image resolution.
The basic design is to point letters from an on-screen keyboard by gaze and select by voluntary blinks. Our keyboard layout is close to the classic 9-key T9 input method keypad used on bar phones. The T9 input method has many advantages: first, it requires less buttons in comparison to the QWERTY keyboard. Less buttons support larger on-screen targets which has a larger fault tolerant space on the screen, allowing for higher selection accuracy. Second, as the 9-button keyboard layout is widely used in bar phones, most people possess the experience that would enable them to easily become familiar with the layout. Also, there are many language models of the T9 input method, which can greatly increase the text entry speed. On a bar phone with a word prediction and completion function, users can at most type in 46 words per minute using the T9 input method~\cite{t9}. Third, the 9 buttons of the T9 input method can be seen in the 9 directions in our system. Buttons 1 to 9 represent left-up, up, right-up, left, middle, right, left-down, down and right-down respectively. Unlike previous designs in which people have to watch the screen to make selections, as people can find the 9 directions by feeling, users can choose the targets without a screen. So, our input method can run both in screen mode and off-screen mode. For off-screen mode, our text entry system can be portable and users can use the system anytime and anywhere. Fig.~\ref{fig:07} shows the overview of the proposed method.

The contribution of this work is threefold.
\begin{enumerate}
\item We collect a number of eye images from 30 participants for making a useful dataset, which is made available to the community to allow further research on this topic.
\item We build a CNN model for 9-direction gaze estimation based on our new dataset. Further, we design several data augmentation methods to increase the robustness of our gaze estimator.
\item We present an input method based on the state of the eyes in each frame of a video sequence. The input method is able to handle false detected eyes and filter noise from involuntary blinks, saccade and momentary false estimation.
\end{enumerate}

\section{Related work}
In order to track the eyes, the first step is to detect the human eyes from the video sequence frame by frame, then to estimate the gaze direction in accordance with the eye detection and calculate the positions on the screen by calibration. There are three popular methods of gaze estimation: model-based gaze estimation, electrooculogram (EOG), and appearance-based gaze estimation~\cite{majaranta2014eye,hansen2010eye}.
The model-based method mainly focuses on analyzing the physical model of human eyes. We can further divide the model-based method into corneal reflection and shape-based methods on the basis of whether external lights are used to irradiate the eyes. When using corneal reflection to detect eye features, infrared lights illuminate the surface of the cornea and multiple cameras are needed to collect eye images~\cite{majaranta2014eye,morimoto2002detecting}. For the shape-based method, gaze directions are inferred directly on the basis of detected eyes shapes, such as corneal center and iris edges~\cite{chen20083d}. A disadvantage of the model-based method is that it requires relatively strong lights and high resolution images.
The EOG method simulates human eyes as a dipole. The cornea is its positive pole and retina is its negative pole~\cite{majaranta2014eye}. If cornea-retina has a stable potential difference, we can detect a steady electric potential field. The eye movement can change dipole orientation, which results in variation of electric potential field. The drawback of EOG is that electrodes are attached around the eye region and EOG is easily influenced by surrounding noise. In addition to that, EOG increases the application cost and is constrained by the applicable environments~\cite{majaranta2014eye}.
The appearance-based method is able to directly use human eye images as the input and can effectively handle low-resolution images. The method learns the features for recognition based on lots of training images, which are then used to predict gaze direction. The model-based and EOG methods were widely used previously and saw success in products, but the appearance-based method was seldom used in application. Even though deep learning has made a huge success in object recognition recently, and the learning-based method attracts numbers of research, the performance of the appearance-based method is still not ideal. This method requires a huge amount of eye image data and there is uncertainty as to whether the estimators could apply to unknown users~\cite{zhang2015appearance}. During the gaze estimation process, the first two methods have an understandable error, because they construct a relationship between extracted features and gaze directions. However, appearance-based gaze estimation builds a relationship between pixel values and gaze directions. The latent learned features used for estimation is unknown. Even though people keep fixed eye direction, the slight variation of illumination may result in different estimation results. Thus, the appearance-based method may produce unstable and discontinuous results during real-time tracking.

There are three popular ways of eye typing. The first method is typing by point. Commonly, an on-screen keyboard is shown and people operate the keyboard by looking at the keys.  Using a QWERTY keyboard ~\cite{istance1996qwer,adjustqwer,gips1996eagleeyes} can achieve a high text input speed, however, it demands accuracy regarding the eye tracker and the error rate can be higher where there is inaccuracy. A hierarchical keyboard~\cite{hansen,hansen2002eye,gips1996eagleeyes} chooses a letter through multiple selections. For example, choosing a letter group then choosing a letter from that letter group. It can decrease the error rate and has a low requirement of devices, but as a single letter is input with multiple selections it has a lower input speed. It is usually used in low-cost systems. Fixation or muscle movement such as blinks are used to make selections. If blinks are used to make selections, involuntary and voluntary blinks must be distinguished. While fixation is used to make selections, the time of fixation is called dwell time. It sets a maximum text input speed and is easily interrupted by blinks and jitter eye movements. The second method is typing by eye-switches. A common way is that an alphabet is shown on the screen and rows or columns are scanned automatically. When the rows or columns where the target letter is located are scanned, users can blink eyes to make a selection as a binary switch~\cite{grauman2003communication,krolak2012eye}. This method is used for people who can not move eyes but can still blink. Coarse eye movement can also be used as switches. For example, looking left is to start while looking right is to select~\cite{majaranta2007text}.
The third way is typing by gestures. I4control system ~\cite{I4} uses eyes to control a cursor moving over an alphabet and blinks are used to make selections. Eyes work like a joystick. Wobbrock ~\cite{wobbrock2007not} uses letter-like gestures to input letters, which uses Edgewrite input method that is used in PDAs, joysticks and trackballs. It is to design an eye gesture for each letter. Dasher system ~\cite{dashereyes} uses continuous pointing gestures with a zooming interface. It used dasher language model to predict the next possible letter. Commonly used vocabulary and sentences can be input with a higher speed and it has lower error than common on-screen QWERTY keyboard. However, it only supports 4 languages.

\section{Dataset and CNN}
\begin{figure*}[t]
\centering
 \includegraphics[width=1\textwidth]{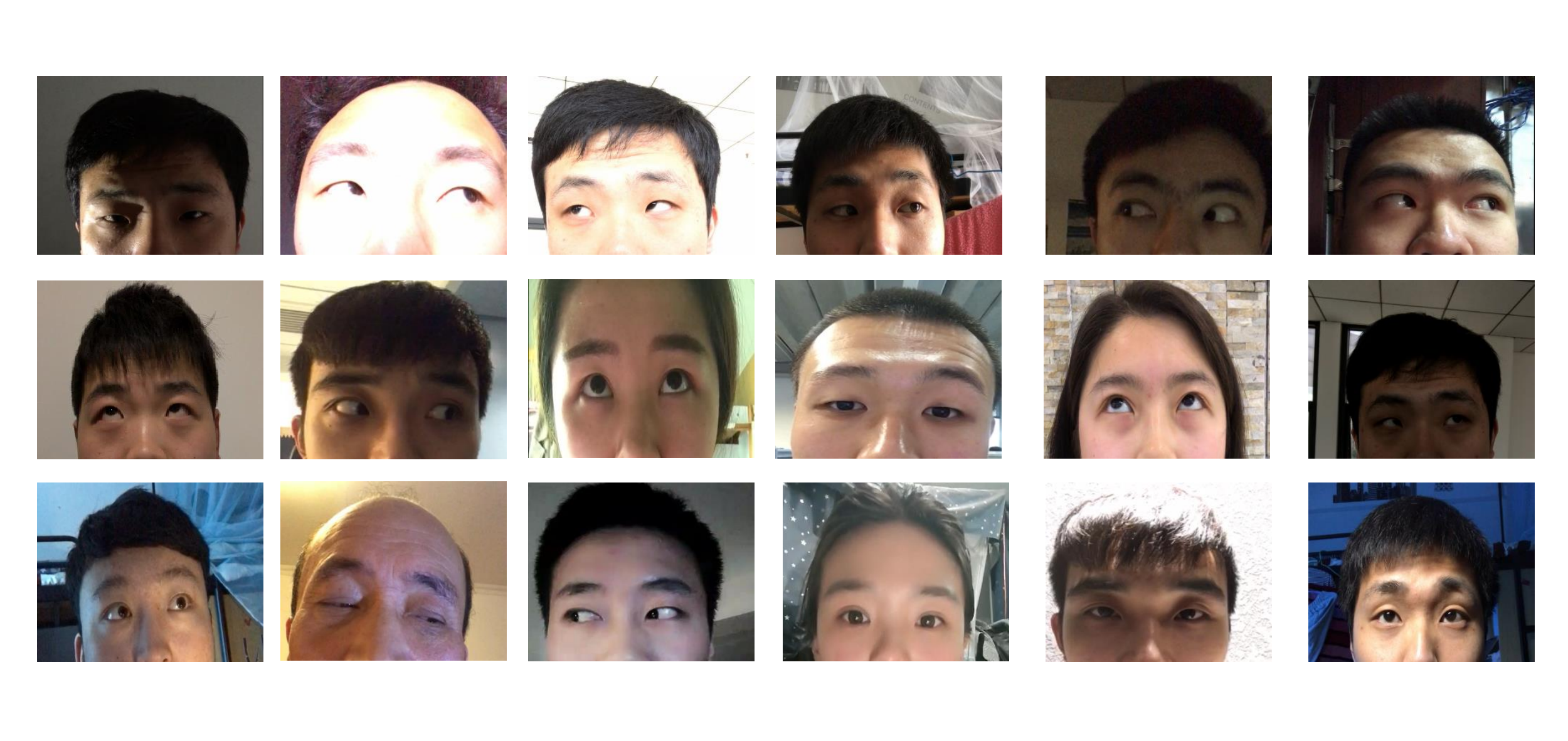}
\caption{Sample images from our dataset}
\vspace{-3mm}
\label{fig:03}
\end{figure*}

\begin{figure}[t]
\centering
 \includegraphics[width=.3\textwidth]{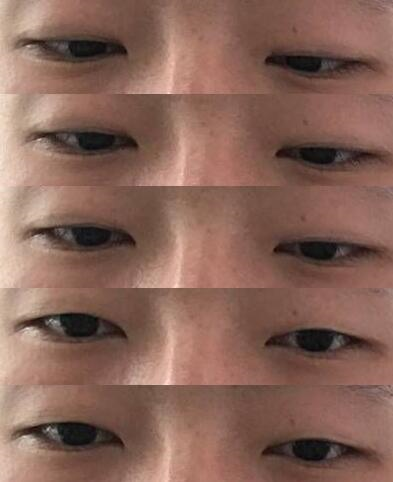}
\caption{A participant looking at different targets locating in the same angle. From top to down, the targets are located 5 cm, 10 cm, 15 cm, 30 cm and 50 cm from eyes}
\label{fig:04}
\end{figure}

In this section, we first introduce a large-scale dataset and then train a CNN for gaze estimation. We divide eye states into 10 categories which consist of 9 directions and an eye-closed state. It took us about 1 month to collect and process eye images from 25 participants. It covers various eye appearance, illumination, locations and time. Eye images are taken in everyday life rather than controlled lavatory conditions. Our collecting devices are mobile phones, webcams and digital cameras. We also implement many data augmentation methods to expand our dataset before and during training to make our gaze estimator more invariant to illumination, skin color, image resolution and slight head rotation. Fig.~\ref{fig:03} shows some sample images from our dataset. We apply cascade double-eye detector using Viola-Jones algorithm~\cite{viola2001rapid} to detect eyes and we use images of both eyes to train and test. Previous appearance based gaze estimation works often take single-eye images as input data, but it may cause problems in some cases. By using the single eye to estimate gaze, a strong assumption is that both eyes are synchronized when looking at a target. However, two eyes are not always synchronized. Fig.~\ref{fig:04} shows a participant looking at targets locating at the same gaze direction but different distances. Even though eyes are looking at the same angle, synchronization of two eyes varies. As is shown, usually the closer a target is to the eyes, the less likely it is for both eyes to be synchronized. Particularly, eyes become crossed when the target moves too close to the middle of both eyes. If single eye is used to estimate gaze, different results may be achieved from two eyes. Another advantage of using images of both eyes is that even though one eye is unclear and impossible to use to estimate gaze due to poor lighting conditions or low resolution, the estimator can still estimate gaze from double eyes. Double-eye appearance based gaze estimation learns features from the eye region and the combinational features of both eyes can be learned by a CNN to better estimate gaze.

\subsection{Data collection and augmentation}
\begin{figure}[t]
\centering
 \includegraphics[width=.5\textwidth]{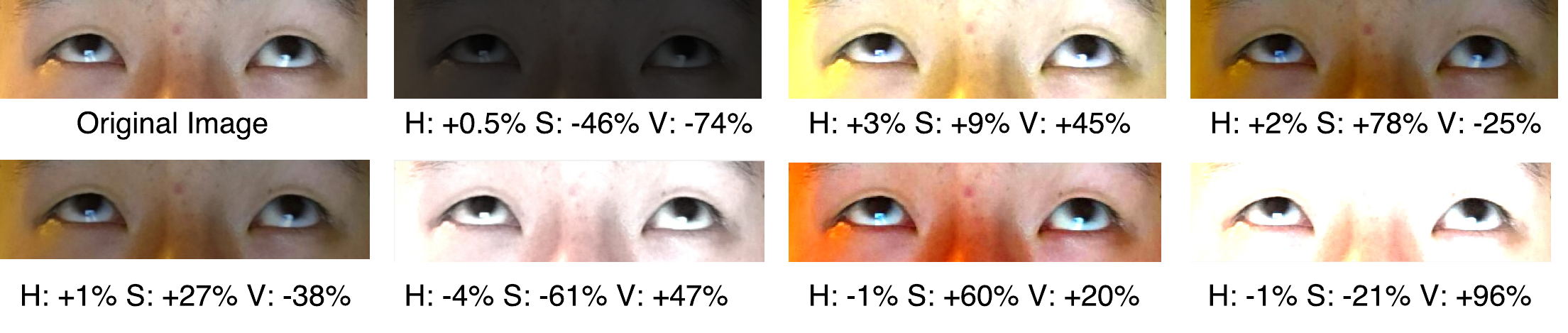}
\caption{Effects on images after the adjustment is made on HSV channels of images}
\label{fig:05}
\end{figure}

First, we recorded 162 video sequences from 25 people under different lighting conditions and each video lasts about 10 to 30 seconds. Two different ways were used to record the videos. The first way is that the camera is 20 cm in front of the face and the camera lens are level with eyes. They were asked to look at 9 directions by feeling and close eyes successively. The second way is that the participants look at 9 targets on a board and close eyes successively. The size of the board is 30 cm $\times$ 50 cm, which is 35 cm far from the face. The camera is put in the center of the board. Camera lens and eyes are on the same level. Each video covers 10 states of the eyes. We used two ways to collect dataset because our input method runs in two modes: screen mode and off-screen mode. In the off-screen mode, the angle of gaze has a wider range while in screen mode, the angle of gaze is limited in the screen region. After all the videos were recorded, we cropped eye images with the eye detector. As we use the eye detector to detect eyes for estimation, the detector knows the best eye region to crop for estimation. Then we manually picked 1 to 3 eye images of a state that look different in each video and record their labels and frame numbers in the video. For the validation set and the test set, we asked participants to take videos at a different time and repeat the same procedures instead of splitting collected data, because this can avoid overfitting caused by similar eye images. We prepared two test sets. One is from people who contribute eye images to the training set. We call them known people. The other one is from new people and we call them unknown people. Even though test images from known people can achieve better results, the test of unknown people can evaluate whether a estimator can be applied to arbitrary users. So far we collected 832 training images, 728 validation images, 501 test images from known people and 434 test images from unknown people. Then we implemented three data augmentation methods to expand our dataset and increase robustness. First, according to the frame numbers we recorded before, we applied some image transformation methods. We rotated the face images by $2.5^{\circ}$ and $-2.5^{\circ}$ and detect and crop the eye region again. We also scaled detected eye regions by 1.2 and 1.5 times and made 5-pixel shifts of the detected regions along 8 directions. Then we saved them as new training images with the same labels. We repeated the transformation to every frame we recorded. This can make the estimator more robust with respect to slight head movement and detected eye regions. Then we manually deleted false detected eye images. We expanded our training set to 74101 eye images. Second, during training, we implemented the method used in~\cite{zhang2015appearance} which is to flip the eye images horizontally with the probability of 0.5 and change the label to the opposite direction, namely swapping labels between left-up and right-up, left and right and left-down and right-down while others remaining the same. Third, to make our estimator robust with respect to illumination, in training process we made a small random adjustment to HSV channels of eye images. By changing the values in HSV channels, we changed the hue, saturation and value level of images. Fig.~\ref{fig:05} shows the results of images after adjustment of HSV channels is made.

\subsection{CNN architecture}
CNN is a powerful kind of neural networks that has a great performance in object recognition. Normally CNN is composed of four parts, including convolutional layers, pooling layers, non-linearity layers, and fully-connected layers. In the following, we will give a brief introduction to them respectively and to our network structure.

In order to apply convolution, we apply convolutional filtering over the image. After sliding a learnable filter over the RGB image and computing the dot product, we get a feature map, which is also called an activation map. The primary purpose of convolution is to extract features from the input image. It can also preserve the spatial relationship between pixels by learning features from input data. Spatial Pooling, also called subsampling or downsampling, reduces the dimensionality of each feature map but retains the most important information. It can be of different types: Max pooling, Average pooling, Sum pooling etc. Max pooling is used in our architecture. We use a 2 $\times$ 2 window sliding over the feature map and keep the largest element within the window. One typical non-linearity operation is called ReLU (Rectified Linear Unit), which is employed in our network. This operation replaces all negative pixel values in the feature map with zero. The reason for using non-linearity operations is that most real-world data is non-linear. Thus this operation allows our network to better recognize objects from the real world. The fully-connected layers consist of fully connected Multi-Layer Perceptrons that use ReLU as the activation functions in our model. The term ¡°Fully Connected¡± implies that every neuron in the previous layer is connected to every neuron in the next layer. Generally, in CNN, convolutional layers, pooling layers, and non-linearity layers are mainly used for feature extraction. The classification is done by the trainable fully-connected layers. We also employ batch normalization layers~\cite{bnlayers} in our model. Batch normalization layers perform normalization for mini-batch in the training process. It can act as a regularizer and works effectively in accelerating training, reducing reliance on weight initialization and avoiding overfitting.

\begin{figure*}[t]
\centering
 \includegraphics[width=1\textwidth]{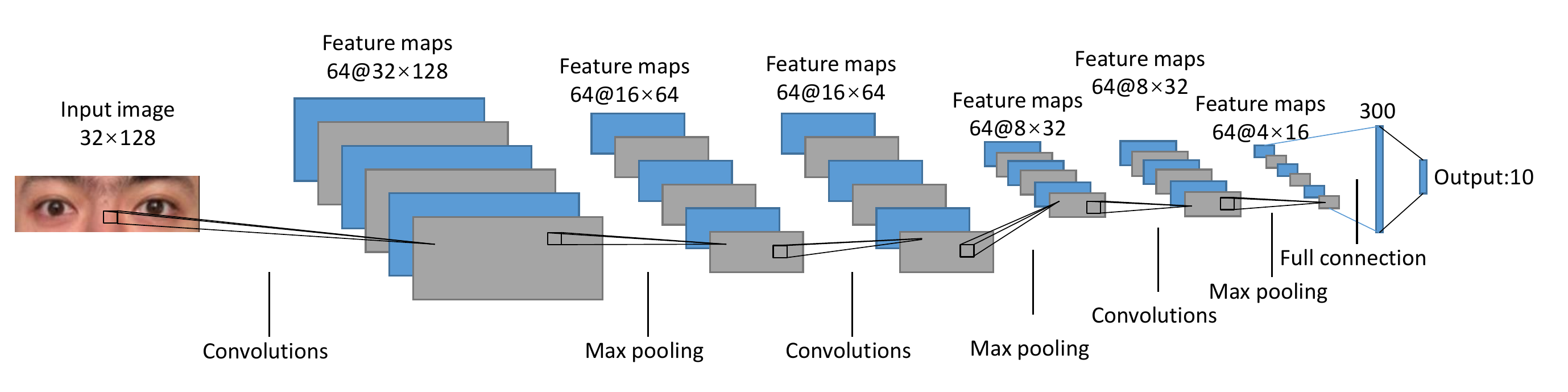}
\caption{The architecture of our CNN model. The CNN consists of 4 blocks. The first three blocks have the same operations which include convolution with 64 3 $\times$ 3 filters, batch normalization, ReLU and max pooling. The fourth block is a fully-connected layer with ReLU as the activation function. The input is an RGB image with the size of 32 $\times$ 128 pixels. The output is the scores of 10 categories.}
\label{fig:06}
\end{figure*}

The architecture of our CNN model is shown in Fig.~\ref{fig:06}. We use RGB images with a size of 32 $\times$ 128 pixels as the input to the network. For the first convolutional layer, the feature size is 3 $\times$ 3 and we pad zeros around the input image. The number of filters is 64. After the convolutional layer, we add a batch normalization layer, followed by ReLU and the max pooling layer. We repeat the same operation on second and third convolutional layers. The fully connected layer has 300 hidden units and each unit has connections to all feature maps of the third convolutional layer. The output of our network is ten categories. We use softmax loss function to calculate the loss for optimization.

\section{Gaze controlled text entry}

In this section, we introduce how we type in words with gaze directions and voluntary blinks. Our general method is pointing by gaze directions and making selections by voluntary blinks. For a video sequence, we need to process each frame. We first detect both eyes from an image and then crop the eye region. After normalization, the eye images are used to estimate states of the eyes. If no eyes are detected in the image, the frame is skipped. If multiple eyes are detected, we eliminate false eyes by the gaze estimator. We choose the detected double eyes with the highest score in the CNN output layer as the real eyes. Then we input text according to the states of the eyes. To make the input system work properly, we need to handle false estimated states, and filter estimated states of saccade and involuntary blinks.

\subsection{Signal analysis}

\begin{figure}[!h]
\centering
 \includegraphics[width=.5\textwidth]{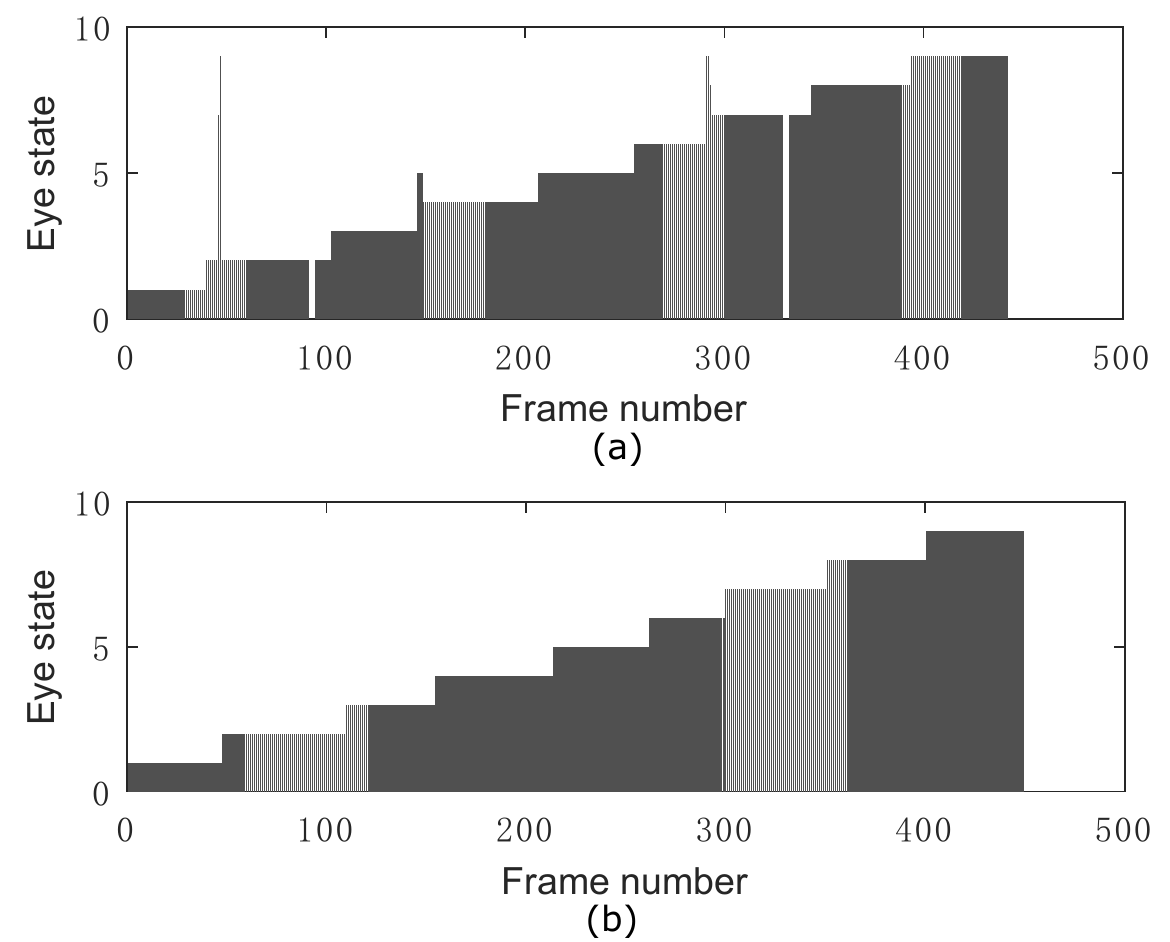}
\caption{The states of the eyes in each frame of the video sequence (29 fps). Numbers 1 to 9 represent the 9 directions and 0 means the state of closed eyes. In the video, the participant moves their eyes from numbers 1 to 9 and then closes their eyes. Natural blinks have occurred during the process. The diagrams (a) and (b) show the results before and after the filter respectively.}
\label{fig:08}
\end{figure}

\begin{figure}[!h]
\centering
 \includegraphics[width=.5\textwidth]{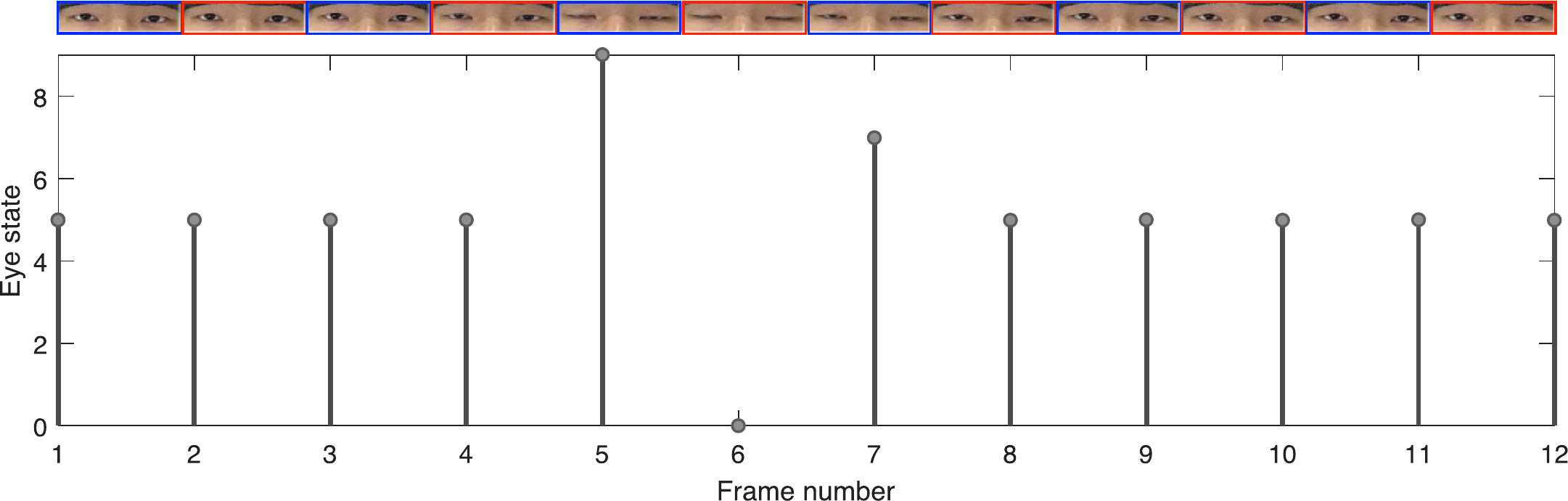}
\caption{A blink process and estimated states of each frame (29 fps).}
\label{fig:09}
\end{figure}

Noise mainly comes from natural blinks, saccade and false estimation. Fig.~\ref{fig:08} (a) shows estimated states of eyes in each frame of a video sequence. As is shown,the signal is stable and accurate during fixation and is unsure during saccade and blinks. Fig.~\ref{fig:09} shows a natural blink process and its estimated states of each frame. The noise lasts 3 to 5 frames which are 0.1-0.17 seconds. So it is important to distinguish voluntary and involuntary blinks and filter noise caused by involuntary blinks, saccade and momentary false estimation to avoid false pointing and selection. Our method is that we set an array recording estimated states in the last 16 frames and the array is updated every time a new frame comes. The earliest state in the array is popped. The most frequent state in the array is set as the current state. So normally only when eyes fixate at a direction for more than 8 frames, the direction can be recognized as the current state, while momentary noise (less than 8 frames) does not influence correct pointing and selections. Fig.~\ref{fig:08} compares the estimated states before and after the filter. As natural blinks usually last less than 5 frames and saccade lasts less than 4 frames, noise caused by them can be filtered. The length of the array is set according to the frame rate and the noise duration. Usually, it is set as more than twice the length of the longest noise duration.

\subsection{Input method}

\begin{figure}[!h]
\centering
 \includegraphics[width=0.48\textwidth]{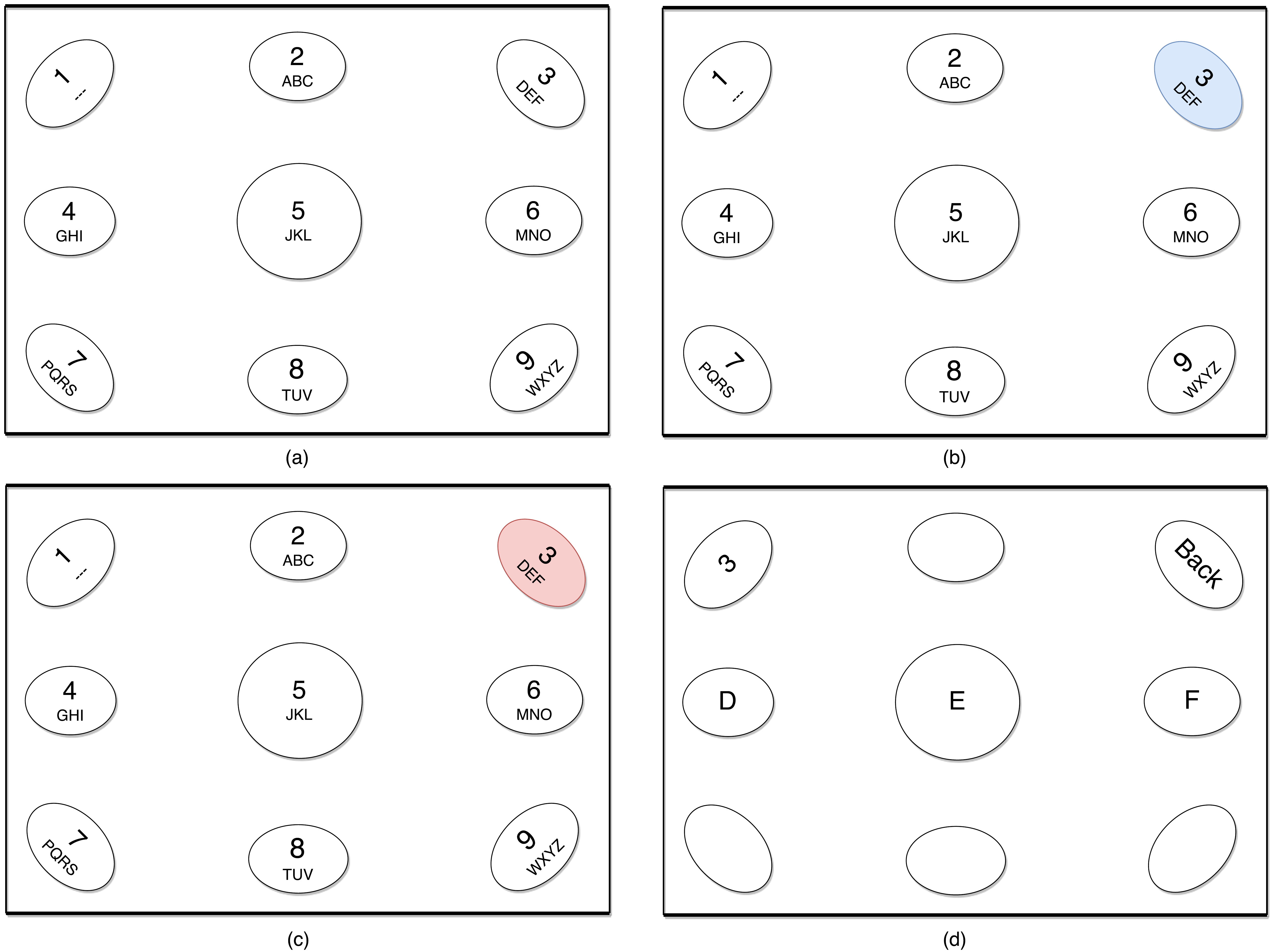}
\caption{Main interface (a), main interface when users look at number 3 (b), main interface when users blink to select number 3 (c) and secondary interface of button 3 (d).}
\label{fig:10}
\end{figure}

Our input method runs in two modes: screen mode and off-screen mode. For screen mode, the camera is table mounted. A keyboard is shown on the screen. For off-screen mode, the camera is head mounted. The camera is right in front of the eyes and users can use our system in any position. In our interface, letters are distributed in buttons 2 to 9 and functional keys are in button 1, which accords with most layouts on mobile phones. After a button is selected by a blink in the main interface, the interface jumps to the secondary interface and then users select the letters or numbers. After a number or a letter is entered, the interface jumps back to the main interface (See Fig.~\ref{fig:10}).

In our text entry system, when gaze moves to a new direction two types of feedbacks will be given to users. In the off-screen mode, the feedback is sound. The system reads out the letters or numbers that the users are looking at. When users close their eyes to make a selection, a `click' sound is played, which means a voluntary blink has been recognized and users can then open their eyes and continue to use. Users do not have to close eyes for a long time to distinguish voluntary and involuntary blinks, which can increase input speed. In screen mode, the feedbacks include sound and real-time visual effects on the screen. The buttons that the users are fixating change color which is synchronized with the sound. With feedbacks, users can know whether the correct direction is estimated and wrong selections can be avoided.

\section{Experiments and discussion}

In this section, we made two test sets from known users and unknown users to evaluate the performance of our CNN model and compare that with other models. We also tested the effectiveness of our data augmentation methods. Comparing the performance of gaze estimation using images of single eye and both eyes is also included in this part. We then invited five users to operate our input method and recorded their mean error rate and text input speed.

\subsection{Gaze estimation evaluation}
We compared the accuracy of our model with LeNet-5 model and Support Vector Machine (SVM). LeNet-5 is a CNN model that has a great success in recognizing handwritten characters. We trained a SVM with a polynomial kernel using HOG features~\cite{hog} of the same training images.  We also experimented to test the performance of our model with input images of single eye. The only difference is that the size of input single-eye image is 32 $\times$ 64 pixels. We cropped images of single eye from our dataset and the estimation of both left and right eyes uses the same model. The test result is shown in Table~\ref{Tbl:Suspiciousness}.
\begin{table*}
\centering
\caption{The top 1 and top 2 accuracy of different models using different test sets and using or not using our data augmentation methods. Our model (SE) denotes our model with single-eye data.}
\label{Tbl:Suspiciousness}
\scalebox{1.0}{
\begin{tabular}{|c||c|c|c|c|c|}
\hline
\multirow{2}{*}{model}  &\multirow{2}{*}{Data Augmentation}& \multicolumn{2}{c|}{Test set of known users} & \multicolumn{2}{c|}{Test set of unknown users} \\
\cline{3-6} & & Top1 Acc. & Top2 Acc. & Top1 Acc. & Top2 Acc. \\  \hline
Our model    & No& 82.44& 95.61 & 78.34 & 95.85    \\
Our model & Yes& 95.01 & 99.80 & 91.71 & 98.39   \\
Lenet-5  & No& 68.46 & 90.02 & 67.28 & 88.94  \\
Lenet-5  & Yes& 80.44 & 94.81 & 81.57 & 93.32  \\
SVM  & No& 70.66 & 87.62 & 70.51 & 89.63  \\
Our model (SE) & Yes & 76.27& 93.31 & 73.15& 92.84   \\ \hline
\end{tabular}
}
\end{table*}

As is shown in the table, our model and data augmentation methods are effective for gaze estimation of known and unknown people. We observed the incorrect estimation images and found that the incorrect estimated direction is usually near the correct direction. Thus, we infer that it is likely to confuse the CNN model to make an incorrect estimation when users watch the edge of two adjacent directions. If we calculate the top 2 error, the error rate will decrease to 0.2\%. This kind of error is acceptable because when using the system to input words, users always watch the middle part of every direction rather than look at the edge of the direction. Moreover, we can watch the results on the screen and listen to the feedback sound at real time. Based on feedbacks, we can make a quick adjustment if the recognized direction is not desired. For example, when the system misunderstands left-up as left, we can adjust gaze closer to up direction.

\begin{figure*}[!htb]
\centering
 \includegraphics[width=1\textwidth]{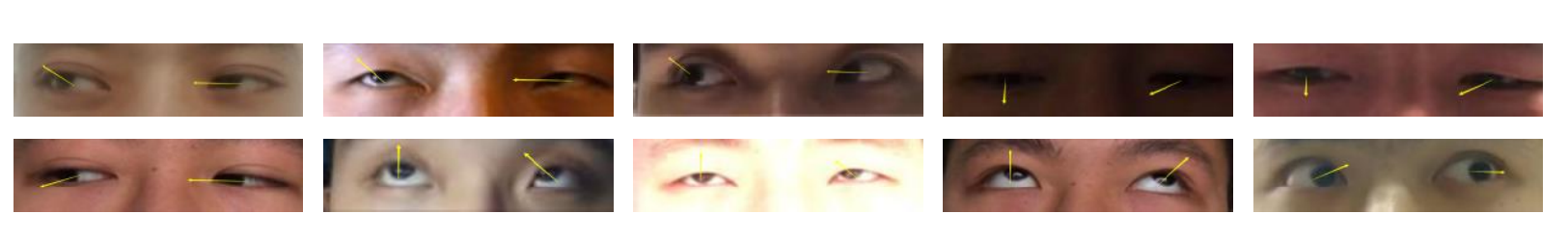}
\caption{Images of both eyes where left and right eyes have different estimated results. The estimated directions of left and right eyes are indicated by arrows on the images. }
\label{fig:11}
\end{figure*}

According to observing on the test results, we find our model with single-eye data has a higher error rate. We examine all false estimated human eye data and find the following three reasons. First, single eye data is more sensitive to slight head movement. Head tilting significantly influences the estimation result. Second, under low resolution and weak lighting conditions, images of single eyes may be too blurry to use for estimation so that even human cannot recognize the direction. Third, left and right eyes are not always synchronized. The estimated results of two eyes are different but both are reasonable. Fig.~\ref{fig:11} shows images of eyes in which left and right eyes have different estimation results.

\subsection{Gaze typing experiments}
We invited 5 participants to test our text entry system. They all have experiences with the T9 input method on bar phones. Before they used our system, we made a calibration for every user. The calibration method is very simple. On screen mode, the screen size is 30 cm $\times$ 40 cm. We used a common webcam. The camera was table mounted. We only required the tester to keep eye level with the center of the screen, then adjusted the distance between the tester and the screen to make sure gaze can be correctly estimated when they look at the nine regions on the screen. We recorded the distance for each user for later use. The calibration took 10 to 20 seconds for each user and the distance between the head and the screen was around 25 to 40 cm. On off-screen mode, we just needed to adjust the distance between the camera and eyes to make sure the camera captures the whole eye region. The distance was around 20 to 30 cm.

\begin{figure}[!h]
\centering
 \includegraphics[width=.5\textwidth]{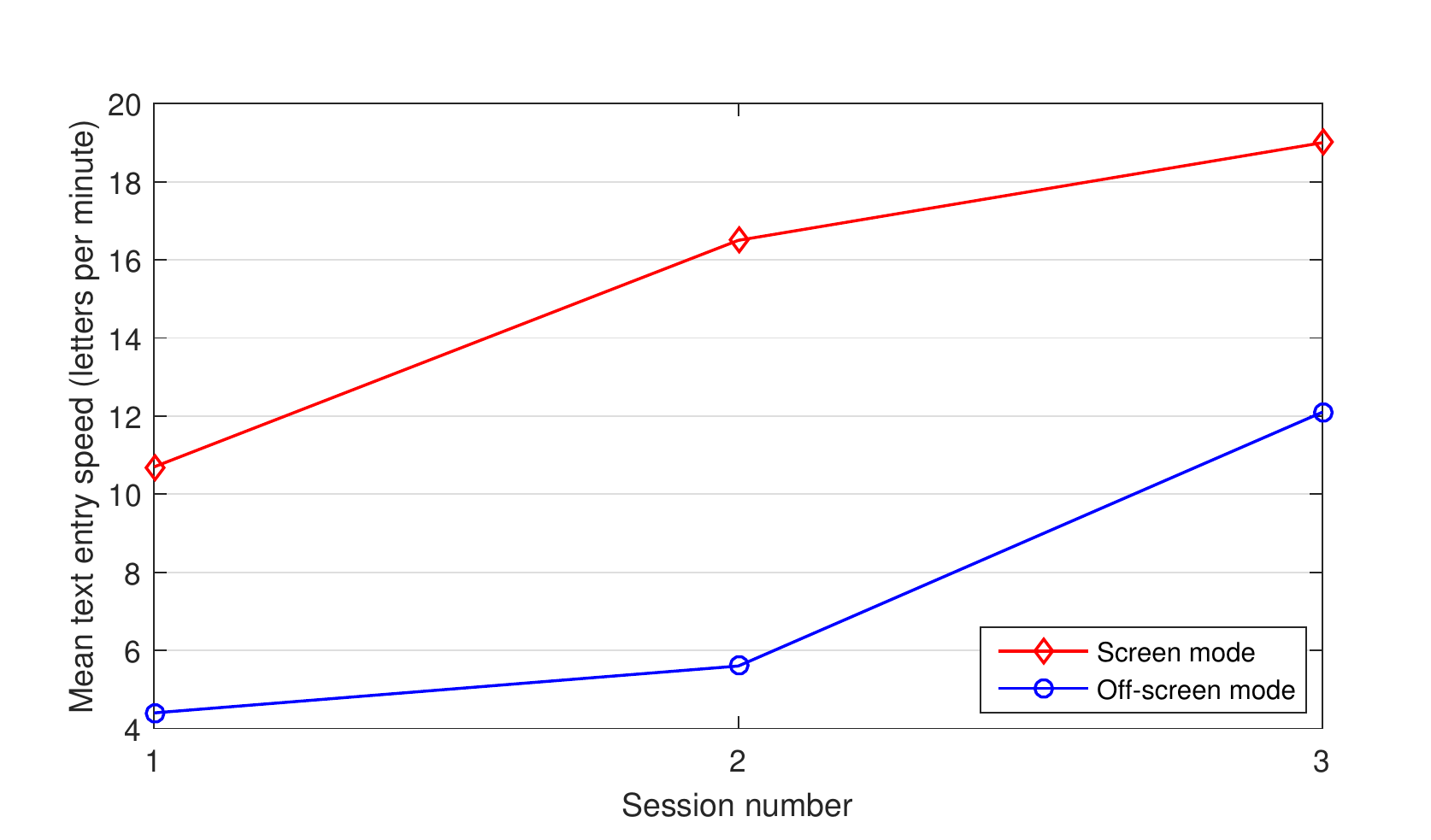}
\caption{Mean text input speed of 5 testers in three sessions }
\label{fig:12}
\end{figure}

\begin{figure}[!h]
\centering
 \includegraphics[width=.5\textwidth]{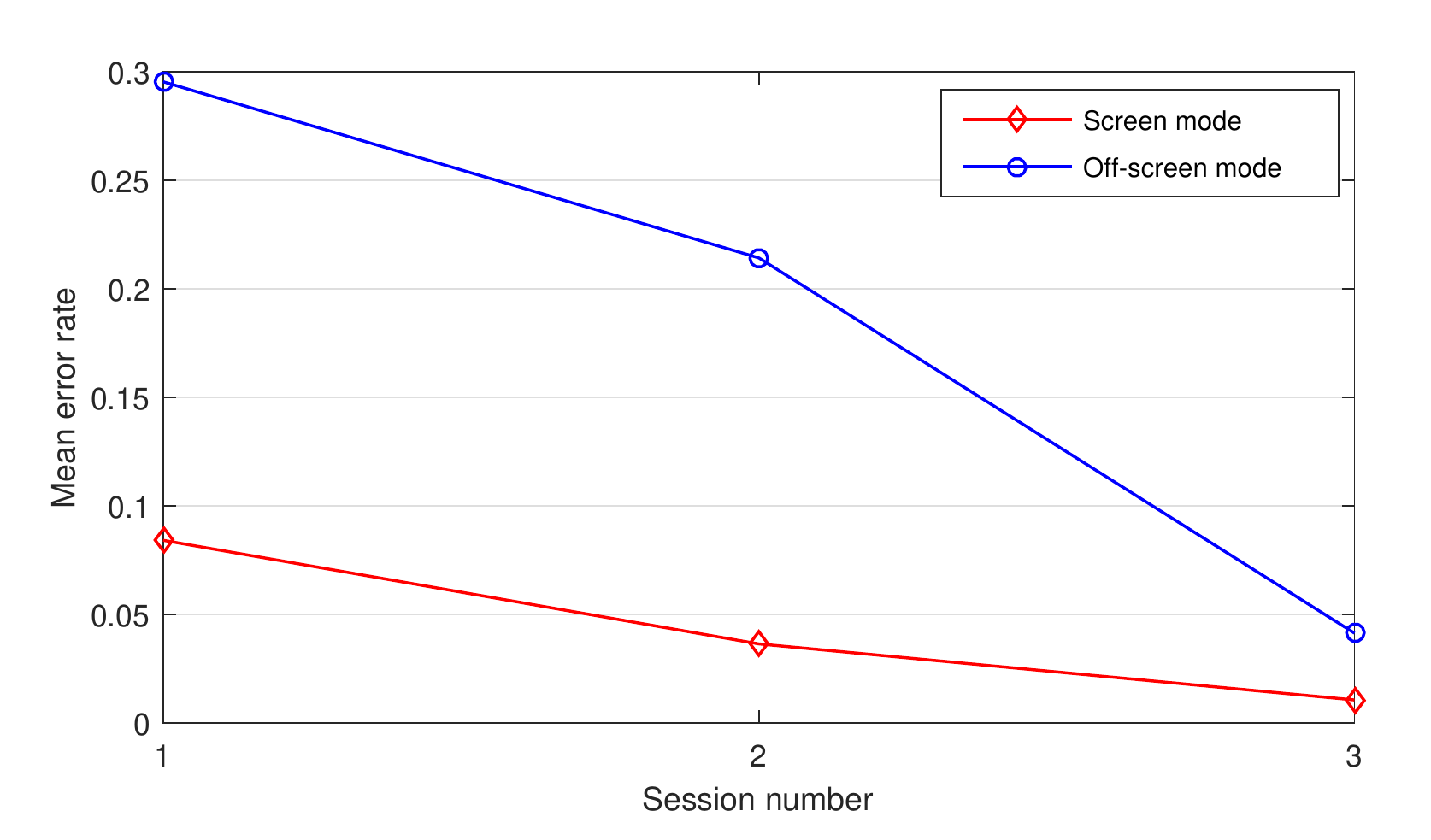}
\caption{Mean error rate of 5 testers in three sessions}
\label{fig:13}
\end{figure}

Our test consisted of three sessions. First, we introduced our system to testers and then they had 10 minutes to familiarize with it before the first test session. There was one hour of training time before the second session where they trained half an hour for each mode. There is a one-day interval between the second and third session. They were required to keep our keyboard layout in mind before they took part in the third test. They were free to use our system for training during the one-day interval. The reason for this arrangement is that we intend to simulate three kinds of users: novices, regular users and experienced users. We test screen mode at first and off-screen mode afterwards in each session. The test duration is 2 minutes for each session. They were asked to type in sample words they knew the spelling of. If testers forget how to spell the words, we will give them support. They were required to type in as many words as possible during 2 minutes. If they make a mistake, they can continue on without a need to delete the incorrect typing. Testers do not need to type a space and spaces will not be counted in total letters. The error rate is the ratio of incorrect letters to the total number of input letters. Fig.~\ref{fig:12} and Fig.~\ref{fig:13} illustrate the mean input speed and error rate of two modes respectively.
We can observe that in both modes there is a great increase in the speed of text input and a decrease in error rates with the increasing of session numbers. The input speed of screen mode is faster than that of off-screen mode for all the testers. Experienced user have a comparable speed of two modes. The fastest speed of screen mode is 20 letters per minute and 18 letters per minute for off-screen mode among all testers.

\subsection{Discussion}

 Testers stated that it was easier to choose letters on screen mode as they were able to find the positions of desired letters on screen easily. On the other hand, they had to spend time recalling the letters¡¯ position when they used the off-screen mode. When they forgot the positions of letters, they had to try directions one by one and rely on the sound feedback to select the correct letter. After they remembered the letters¡¯ position, the speed of input increased greatly. Incorrect text input was all due to users moving their eyes before they blink to choose the correct letter. We can also see that by session three users were able to accurately type in words letter by letter with very low error rate. When people use T9 input method on bar phones, they do not need to type each letter of a word because on bar phones, language models are used to help users type in words efficiently. Users only need to choose the letter groups. For example, to type in 'hello', they only need to select buttons 4, 3, 5, 5 and 6 successively where letters `h', `e', `l', `l' and `o' are located respectively and the language model can predict the desired words. So only one selection is needed to type in a letter on average. As shown in our experiments, our input method can reach up to 20 letters per minute which means we make 40 selections. So if language model is introduced in our input method, text entry speed can double and can reach up to 40 letters per minute theoretically. Also, there are many language models based on 9-key T9 text input keyboard on bar phones so our vision based text input method is expected to support efficient text input of more languages.

\section{Conclusion}
We have designed the vision based text entry method that users can input text by eye movement and blinks. We design the system by solving both eye tracking problems and text input problems. We selected the common bar phones¡¯ 9-key T9 keyboard layout as the interface. A large scale dataset was made which was collected by using several capturing devices and covers different lighting conditions, locations and time. We used images of both eyes to avoid single eye images' synchronization issues, which increased the estimation accuracy. For applications purposes, we define nine gaze directions and train a CNN model for gaze estimation which can precisely estimate known people's and unknown people's gaze. Data augmentation methods we use provide robustness and increase the estimation accuracy. Our text input method utilizes low cost devices, simplifies the calibration steps and filters noise caused by saccade, natural blinks and momentary false estimation during the operation process. Our input method is able to run in two modes: off-screen mode and screen mode. On off-screen mode, users can arbitrarily move and the device is portable. The text input speed of screen mode can reach 20 letters per minute and off-screen mode can reach 18 letters per minute with low error rate. Our future work is to introduce language model and add word prediction and completion functions to our current system and to train a head pose invariant gaze estimator.



\bibliographystyle{IEEEtran}
\bibliography{bib_file}

\begin{thebibliography}{10}
\providecommand{\url}[1]{#1}
\csname url@samestyle\endcsname
\providecommand{\newblock}{\relax}
\providecommand{\bibinfo}[2]{#2}
\providecommand{\BIBentrySTDinterwordspacing}{\spaceskip=0pt\relax}
\providecommand{\BIBentryALTinterwordstretchfactor}{4}
\providecommand{\BIBentryALTinterwordspacing}{\spaceskip=\fontdimen2\font plus
\BIBentryALTinterwordstretchfactor\fontdimen3\font minus
  \fontdimen4\font\relax}
\providecommand{\BIBforeignlanguage}[2]{{%
\expandafter\ifx\csname l@#1\endcsname\relax
\typeout{** WARNING: IEEEtran.bst: No hyphenation pattern has been}%
\typeout{** loaded for the language `#1'. Using the pattern for}%
\typeout{** the default language instead.}%
\else
\language=\csname l@#1\endcsname
\fi
#2}}
\providecommand{\BIBdecl}{\relax}
\BIBdecl

\bibitem{twentyyear}
P.~Majaranta and K.-J. R{\"a}ih{\"a}, ``Twenty years of eye typing: systems and
  design issues,'' in \emph{Proceedings of the 2002 symposium on Eye tracking
  research \& applications}.\hskip 1em plus 0.5em minus 0.4em\relax ACM, 2002,
  pp. 15--22.

\bibitem{visualfield1}
G.~Dagnelie, ``Visual prosthetics: Physiology, bioengineering,''
  \emph{Rehabilitation}, 2011.

\bibitem{visualfield2}
K.~K.~C. Dohse, \emph{Effects of field of view and stereo graphics on memory in
  immersive command and control}.\hskip 1em plus 0.5em minus 0.4em\relax Iowa
  State University, 2007.

\bibitem{hansen2010eye}
D.~W. Hansen and Q.~Ji, ``In the eye of the beholder: A survey of models for
  eyes and gaze,'' \emph{IEEE transactions on pattern analysis and machine
  intelligence}, vol.~32, no.~3, pp. 478--500, 2010.

\bibitem{krafka2016eye}
K.~Krafka, A.~Khosla, P.~Kellnhofer, H.~Kannan, S.~Bhandarkar, W.~Matusik, and
  A.~Torralba, ``Eye tracking for everyone,'' in \emph{Proceedings of the IEEE
  Conference on Computer Vision and Pattern Recognition}, 2016, pp. 2176--2184.

\bibitem{t9}
M.~Silfverberg, I.~S. MacKenzie, and P.~Korhonen, ``Predicting text entry speed
  on mobile phones,'' in \emph{Proceedings of the SIGCHI conference on Human
  Factors in Computing Systems}.\hskip 1em plus 0.5em minus 0.4em\relax ACM,
  2000, pp. 9--16.

\bibitem{majaranta2014eye}
P.~Majaranta and A.~Bulling, ``Eye tracking and eye-based human--computer
  interaction,'' in \emph{Advances in physiological computing}.\hskip 1em plus
  0.5em minus 0.4em\relax Springer, 2014, pp. 39--65.

\bibitem{morimoto2002detecting}
C.~H. Morimoto, A.~Amir, and M.~Flickner, ``Detecting eye position and gaze
  from a single camera and 2 light sources,'' in \emph{Pattern Recognition,
  2002. Proceedings. 16th International Conference on}, vol.~4.\hskip 1em plus
  0.5em minus 0.4em\relax IEEE, 2002, pp. 314--317.

\bibitem{chen20083d}
J.~Chen and Q.~Ji, ``3d gaze estimation with a single camera without ir
  illumination,'' in \emph{Pattern Recognition, 2008. ICPR 2008. 19th
  International Conference on}.\hskip 1em plus 0.5em minus 0.4em\relax IEEE,
  2008, pp. 1--4.

\bibitem{zhang2015appearance}
X.~Zhang, Y.~Sugano, M.~Fritz, and A.~Bulling, ``Appearance-based gaze
  estimation in the wild,'' in \emph{Proceedings of the IEEE Conference on
  Computer Vision and Pattern Recognition}, 2015, pp. 4511--4520.

\bibitem{istance1996qwer}
H.~O. Istance, C.~Spinner, and P.~A. Howarth, ``Providing motor impaired users
  with access to standard graphical user interface (gui) software via eye-based
  interaction,'' in \emph{Proceedings of the 1st European Conference on
  Disability, Virtual Reality and Associated Technologies (ECDVRAT¡¯96)}, 1996.

\bibitem{adjustqwer}
P.~Majaranta, U.-K. Ahola, and O.~{\v{S}}pakov, ``Fast gaze typing with an
  adjustable dwell time,'' in \emph{Proceedings of the SIGCHI Conference on
  Human Factors in Computing Systems}.\hskip 1em plus 0.5em minus 0.4em\relax
  ACM, 2009, pp. 357--360.

\bibitem{gips1996eagleeyes}
J.~Gips and P.~Olivieri, ``Eagleeyes: An eye control system for persons with
  disabilities,'' in \emph{The Eleventh International Conference on Technology
  and Persons with Disabilities}, 1996, pp. 1--15.

\bibitem{hansen}
T.~E. Hutchinson, K.~P. White, W.~N. Martin, K.~C. Reichert, and L.~A. Frey,
  ``Human-computer interaction using eye-gaze input,'' \emph{IEEE Transactions
  on systems, man, and cybernetics}, vol.~19, no.~6, pp. 1527--1534, 1989.

\bibitem{hansen2002eye}
D.~W. Hansen, J.~P. Hansen, M.~Nielsen, A.~S. Johansen, and M.~B. Stegmann,
  ``Eye typing using markov and active appearance models,'' in
  \emph{Applications of Computer Vision, 2002.(WACV 2002). Proceedings. Sixth
  IEEE Workshop on}.\hskip 1em plus 0.5em minus 0.4em\relax IEEE, 2002, pp.
  132--136.

\bibitem{grauman2003communication}
K.~Grauman, M.~Betke, J.~Lombardi, J.~Gips, and G.~R. Bradski, ``Communication
  via eye blinks and eyebrow raises: Video-based human-computer interfaces,''
  \emph{Universal Access in the Information Society}, vol.~2, no.~4, pp.
  359--373, 2003.

\bibitem{krolak2012eye}
A.~Kr{\'o}lak and P.~Strumi{\l}{\l}o, ``Eye-blink detection system for
  human--computer interaction,'' \emph{Universal Access in the Information
  Society}, vol.~11, no.~4, pp. 409--419, 2012.

\bibitem{majaranta2007text}
P.~Majaranta and K.-J. R{\"a}ih{\"a}, ``Text entry by gaze: Utilizing
  eye-tracking,'' \emph{Text entry systems: Mobility, accessibility,
  universality}, pp. 175--187, 2007.

\bibitem{I4}
M.~Fejtova, J.~Fejt, and L.~Lhotska, ``Controlling a pc by eye movements: The
  memrec project,'' \emph{Computers Helping People with Special Needs}, pp.
  623--623, 2004.

\bibitem{wobbrock2007not}
J.~O. Wobbrock, J.~Rubinstein, M.~Sawyer, and A.~T. Duchowski, ``Not typing but
  writing: Eye-based text entry using letter-like gestures,'' in
  \emph{Proceedings of The Conference on Communications by Gaze Interaction
  (COGAIN)}, 2007, pp. 61--64.

\bibitem{dashereyes}
O.~Tuisku, P.~Majaranta, P.~Isokoski, and K.-J. R{\"a}ih{\"a}, ``Now dasher!
  dash away!: longitudinal study of fast text entry by eye gaze,'' in
  \emph{Proceedings of the 2008 symposium on Eye tracking research \&
  applications}.\hskip 1em plus 0.5em minus 0.4em\relax ACM, 2008, pp. 19--26.

\bibitem{viola2001rapid}
P.~Viola and M.~Jones, ``Rapid object detection using a boosted cascade of
  simple features,'' in \emph{Computer Vision and Pattern Recognition, 2001.
  CVPR 2001. Proceedings of the 2001 IEEE Computer Society Conference on},
  vol.~1.\hskip 1em plus 0.5em minus 0.4em\relax IEEE, 2001, pp. I--I.

\bibitem{bnlayers}
S.~Ioffe and C.~Szegedy, ``Batch normalization: Accelerating deep network
  training by reducing internal covariate shift,'' \emph{arXiv preprint
  arXiv:1502.03167}, 2015.

\bibitem{hog}
N.~Dalal and B.~Triggs, ``Histograms of oriented gradients for human
  detection,'' in \emph{Computer Vision and Pattern Recognition, 2005. CVPR
  2005. IEEE Computer Society Conference on}, vol.~1.\hskip 1em plus 0.5em
  minus 0.4em\relax IEEE, 2005, pp. 886--893.

\end{thebibliography}
%




\end{document}